\documentclass[letterpaper]{article}
\usepackage{aaai}
\usepackage{times, helvet, courier}
\usepackage{amsmath, amsthm, amssymb, amsfonts}
\usepackage{subfigure, dsfont, graphicx}

\newcommand*\bell{\ensuremath{\boldsymbol\ell}}

\newcommand{\x}{{\mathbf x}}

\newcommand{\wt}{\mathbf{w}}

\newcommand{\bbO}{{\mathds 1}}

\newcommand{\cK}{{\mathcal K}}

\newcommand{\dd}{{\partial}}

\newtheorem{thm}{Theorem}

\newtheorem{eg}{Example}
\newtheorem{defn}{Definition}

\newtheorem{algol}{Algorithm}

\newenvironment{proofsketch}{\vspace{4mm}\noindent{\emph{Proof Sketch.}}}%
        {\hspace*{\fill}$\Box$\par\vspace{4mm}}

\begin{document}

\title{Kickback cuts Backprop's red-tape: \\Biologically plausible credit assignment in neural networks}
\author{David Balduzzi\\
\texttt{david.balduzzi@vuw.ac.nz}\\
Victoria University of Wellington
\And
Hastagiri Vanchinathan\\
\texttt{hastagiri@inf.ethz.ch}\\
ETH Zurich
\And
Joachim Buhmann\\
\texttt{jbuhmann@inf.ethz.ch}\\
ETH Zurich
}

\maketitle

\begin{abstract}
	Error backpropagation is an extremely effective algorithm for assigning credit in artificial neural networks. However, weight updates under Backprop depend on lengthy recursive computations and require separate output and error messages -- features not shared by biological neurons, that are perhaps unnecessary. In this paper, we revisit Backprop and the credit assignment problem.
	
	We first decompose Backprop into a collection of interacting learning algorithms; provide regret bounds on the performance of these sub-algorithms; and factorize Backprop's error signals. Using these results, we derive a new credit assignment algorithm for nonparametric regression, Kickback, that is significantly simpler than Backprop. Finally, we provide a sufficient condition for Kickback to follow error gradients, and show that Kickback matches Backprop's performance on real-world regression benchmarks.	
\end{abstract}

\section{Introduction}

The discovery of error backpropagation was hailed as a breakthrough because it solved the main problem of distributed learning -- the spatial credit assignment problem \cite{werbos:74,rumelhart:86}. Decades later, Backprop is the workhorse underlying most deep learning algorithms, and a major component of the state-of-the-art in supervised learning. 

Since Backprop's introduction, there has been tremendous progress improving the performance of neural networks. An enormous amount of effort has been expended exploring the effects of: the activation functions of nodes; network architectures (e.g. number of layers and number of nodes); regularizers such as dropout \cite{srivastava:14}; modifications to accelerate gradient descent; and unsupervised methods for pre-training to find better local optima.

However, it was known from the start that Backprop is not biologically plausible \cite{crick:89}. Implementing Backprop requires that neurons produce two distinct signals -- outputs and errors  -- whereas only one has been observed in cortex \cite{lamme:00,roelfsema:05}. 

It is therefore remarkable that almost no attempts have been made to rethink the core algorithm -- backpropagation -- and the problem that it solves -- credit assignment. This paper revisits the credit assignment problem and takes a fresh look at the signaling architecture that underlies Backprop. 

\paragraph{Outline.}
Our starting point is to decompose Backprop into local learning algorithms, Theorem~\ref{t:Backprop}. Nodes under Backprop are modeled as agents that minimize their losses. Backprop ensures that nodes cooperate, towards the shared goal of minimizing the output layer's error, by gluing together their loss functions using recursively computed error signals. 

Reformulating Backprop as local learners immediately suggests modifying the signaling architecture (the glue) whilst keeping the learners. In this paper, we aim to simplify Backprop's error signals.

Theorem~\ref{t:regret} lays the groundwork, by providing a regret bound for local learners that holds for any scalar feedback -- and not just the error signals used by Backprop. 

The next step is to show that, when a neural network has 1-dimensional outputs (e.g. nonparametric regression), Backprop's error signals factorize into two components, Theorem~\ref{t:factorization}. The first component is a scalar error computed at the output layer that is analogous to a neuromodulatory signal; the second is a complicated sum over paths to the output layer that has no biological analog.

Our proposed algorithm, \emph{Kickback}, modifies Backprop by truncating the second component. Kickback is \emph{not} gradient descent on the output error. Nevertheless, Theorem~\ref{t:coherence} provides a simple sufficient condition, coherence, for Kickback to follow the error gradient.

It turns out that many of the components of Kickback have close neurophysiological analogs. We discuss Kickback's biological significance by relating it to a recently developed, discrete-time model neuron \cite{bb:12}.  

Finally, we present experiments demonstrating that Kickback matches Backprop's performance on standard benchmark datasets.

\paragraph{Synopsis.}
Our contribution is twofold. Firstly, we provide a series of simple, fundamental theorems on Backprop, one of the most heavily used learning algorithms. In particular, Theorem~\ref{t:Backprop} suggests that ideas from multi-agent learning and mechanism design have a role to play in deep learning.

Secondly, we propose Kickback, a stripped-down variant of Backprop that simultaneously performs well and ties in nicely with the signaling architecture of cortical neurons.

\paragraph{Related work.}
\label{s:related}

The idea of building learning algorithms out of individual learning agents dates back to at least \cite{selfridge:58}. More recent approaches include REINFORCE \cite{williams:92}, the hedonistic neurons in \cite{seung:03}, and the neurons modeled using online learning in \cite{hu:13}. None of these approaches have led to algorithms that are competitive on benchmarks.

The algorithm closest to Kickback is attention-gated reinforcement learning (AGREL), which also eliminates the error signals from Backprop \cite{roelfsema:05}. AGREL and Kickback are analogous at a high level, however the details differ markedly. In terms of results, the main differences are as follows. Firstly, we implement Kickback for networks with 2 and 3 hidden layers; whereas AGREL was only implemented for 1 hidden layer. Indeed, as discussed in \cite{roelfsema:05}, extending AGREL to  multiple hidden layers is problematic. Secondly, AGREL achieved comparable performance to Backprop on toy datasets: XOR, counting, and a mine detection dataset containing $\pm200$ inputs; whereas Kickback matches Backprop on much larger, real-world nonparametric regression problems. Finally, AGREL converges $1.5$ to $10$ times slower than Backprop, whereas Kickback's convergence is essentially identical to Backprop.

\section{Error Backpropagation}

Recent work has shown that using rectilinear functions instead of sigmoids can significantly improve the performance of neural networks. We restrict to rectifiers because they perform well empirically \cite{jarrett:09,nair:10,glorot:11,krizhevsky:12,zeiler:13,dahl:13,maas:13}, are more realistic models of cortical neurons than sigmoid units \cite{glorot:11}, and are universal function approximators \cite{leshno:93}.

Denote the positive and negative rectifiers by $P(a):=\max(0,a)$ and $N(a):=-\max(0,a)$ respectively. Rectifiers are continuous everywhere and differentiable everywhere except at 0. The subgradients are:
\begin{equation*}
	\nabla P(a) := \begin{cases}
		1 & a > 0\\
		0 & \text{else}
	\end{cases}	
	\quad\quad\quad
	\nabla N(a) := \begin{cases}
		-1 & a > 0\\
		0 & \text{else.}
	\end{cases}	
\end{equation*}

Let $S(a)$ denote either $P(a)$ or $N(a)$; the notation is useful when discussing positive and negative rectifiers simultaneously. Similarly, let $\bbO$ denote either subgradient. The subgradient $\bbO$ acts as a \emph{signed} indicator function.

The output of node $j$ is $S_{\wt_j}(\x) := S(\langle\wt_j,\x\rangle)$. We say that node $j$ fires if $\langle\wt_j,\x\rangle>0$; the firing rate is $|S_{\wt_j}(\x)|$.

\paragraph{From global to local learning.}

Under Backprop the entire neural network optimizes a single objective function using  gradient descent on the network's error. The partial derivatives with respect to weights are computed via the chain rule.

In more detail, suppose a neural network has error function $E(\x,y)$ that depends on the output layer $\x_o$ and labels $y$. Backprop recursively updates weight vectors using the chain rule. For nodes in the output layer, $\delta_o:= \frac{\partial E}{\partial x_o}$. For hidden node $j$, the error signal is derived via
\begin{equation}
	\label{e:bp}
	\delta_j 
	:= \sum_{\{k|j\rightarrow k\}}w_{jk}\bbO_k \delta_k.
\end{equation}

Our first result is that, when the hidden nodes are rectilinear, Backprop decomposes into many interacting learning algorithms that maximize local objective functions. 

Consider the following setup. 

\begin{defn}[rectilinear loss]\label{d:rl}
	A node with a rectilinear activation function $S_{\wt}(\bullet)$ receives input $\x$ and incurs \textbf{rectilinear loss}
	\begin{equation*}
		\bell_{RL}(\wt,\x,\varphi) := \varphi\cdot S_{\wt}(\x)
		= \begin{cases}
			\pm\varphi\cdot \langle\wt,\x\rangle & \text{if }\langle\wt,\x\rangle>0\\
			0 & \text{else}
		\end{cases}
	\end{equation*}	
	that depends on an externally provided scalar $\varphi$. 
\end{defn}
If the node fires then the rectilinear loss is the linear loss $\bell_{L}(\wt,\varphi\cdot \x):=\langle\wt,\varphi\cdot \x\rangle$, which has been extensively analyzed in online learning \cite{cesa:06}. If the node does not fire then the rectilinear loss is zero.

\begin{thm}[Backprop decomposes into local learners]\label{t:Backprop}
	
	The weight updates induced by Backprop on rectilinear hidden node $j$ are the same as gradient descent on the rectilinear loss:
	\begin{equation*}
		\nabla_{\wt_j}\bell_{RL}(\wt_j,\x,\delta_j) = \nabla_{\wt_j} E(\x_o,y).
	\end{equation*}
\end{thm}

The rectilinear loss resembles the hinge loss. However, it is not convex since, even if the node has a positive rectifier, $\varphi$ is not necessarily positive.

\begin{proofsketch}
	Let $a_j=\langle\wt_j,\x\rangle$ and $x_j=S(a_j)$.
	Weight updates under Backprop are
	\begin{equation*}
		\Delta w_{ij} \propto -\frac{\dd E}{\dd w_{ij}} 
		= -\frac{\dd E}{\dd a_j}\frac{\dd a_j}{\dd w_{ij}}
		= -\delta_j \cdot x_i\cdot\bbO_{j}.
	\end{equation*}
	Weight updates for gradient descent on the rectilinear loss are
	\begin{equation}
		\label{e:grad1}
		\Delta w_{ij} \propto - \frac{\dd \bell_{RL}}{\dd w_{ij}} = - \varphi\cdot x_i \cdot \bbO_j.
	\end{equation}
	Substituting $\varphi \leftarrow \delta_j$ yields the theorem.
\end{proofsketch}

Backprop is thus a collection of local optimizations glued together by the recursively computed error signals. 

\paragraph{A regret bound.}
Since the rectilinear loss has not been previously studied, our second result is a guarantee on the predictive performance of the local learners.

\begin{thm}[regret bound for local learners]\label{t:regret}	
	Suppose that weights are projected into a compact convex set $\cK$ at each time step.	
	Let $F:=\{t \,|\, S_{\wt^{t}}(\x^{t})>0\}$ denote the time-points when the node fired. 

	The following guarantee holds for any sequence of inputs and scalar feedback when $|F| \geq 1$:
	\begin{gather*}
		\frac{1}{|F|}\left[\sum_{t\in F} \bell_{RL}(\wt^{t},\x^{t},\varphi^t) 
		- \inf_{\wt\in\cK}\sum_{t \in F}\bell_{RL}(\wt,\x^{t},\varphi^t)\right]
		\\
		\leq \sqrt{\frac{8 D E}{|F|}}
	\end{gather*}
	where $D= \max_{t\in F}\left\{ \|\varphi^{t}\cdot \x^{t}\|_2^2\right\}$ and $E=\max_{\wt\in\cK}\|\wt\|_2^2-\|\wt_1\|_2^2$.
\end{thm}

Theorem~\ref{t:regret} shows that the loss incurred by rectifiers on \emph{the inputs that cause them to fire} converges towards the loss of the best weight-vector in hindsight. The theorem is shown for hard constraints (i.e. projecting into $\cK$); similar results hold for convex regularizers.

The result holds for arbitrary sequences of inputs and feedbacks, including adversarial. It is therefore more realistic than the standard \emph{i.i.d.} assumption. Indeed, even if a network's inputs are \emph{i.i.d.}, the inputs to nodes in deeper layers are not -- due to weight-updates within the network. 

\begin{proofsketch}
	Standard results on online learning do not directly apply, since the rectilinear loss is not convex. To adapt these results, observe that, by \eqref{e:grad1}, nodes only learn from the inputs that cause them to fire.

	Clearly, $S_{\wt^{t}}(\x^{t})=\langle\wt^{t},\x^{t}\rangle$ for all $t\in F$. That is, a node's output is linear on the inputs for which it fires. Further, the rectilinear loss is linear on $F$. The theorem follows from a well-known result on gradient descent for the linear loss, see \cite{hazan:12}.
\end{proofsketch}

Theorem~\ref{t:regret} is not restricted to Backprop's error signals; it holds for any sequence of scalars $\{\varphi^t\}$. This suggests exploring alternate ways of gluing together local learners.

\section{Kickback: truncated error backpropagation}
\label{s:kbbp}

\begin{figure*}
	\begin{center}
		{\includegraphics[width=.9\textwidth]{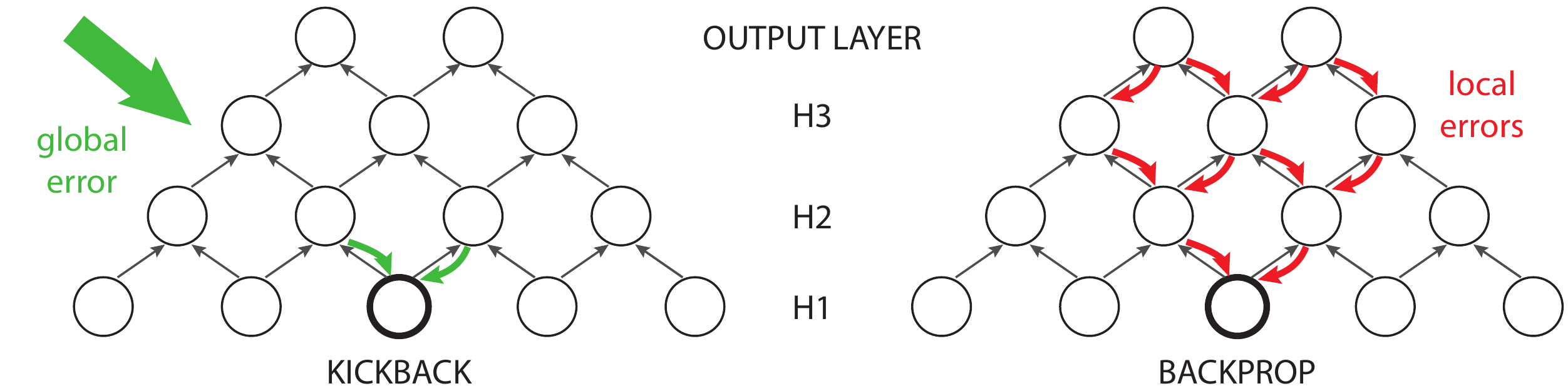}
		\label{f:bp}}
		\caption{\textbf{Schematic comparison of Kickback and Backprop.}
		Black arrows represent feedforward conenctivity. Colored arrows depict paths used to compute the bold node's feedback under each algorithm.
		}
	\end{center}
\end{figure*}

Backprop has two unfortunate properties. Firstly, the error signals $\delta_j$ are computationally expensive: they depend on the activity and weights of all downstream nodes in the network. Secondly, nodes produce two distinct signals: outputs that are fed forward and errors that are fed back. In contrast, cortical neurons communicate with only one signal type, spikes, which are sent in all directions. This suggests that it may be possible to make do with less. 

Viewed from a distance, Backprop is a single distributed optimization, performing gradient descent on the network's error. Zooming in, via Theorem~\ref{t:Backprop}, reveals that Backprop is a collection of local learners \emph{glued together} by recursively computed error signals. We thus have a framework for experimenting with alternate feedback signals \cite{balduzzi:14cpm}.

\emph{Kickback} takes the same local learners as Backprop but weakens the glue that binds them, thereby reducing communication complexity and increasing biological plausibility.

\paragraph{Factorizing Backprop's error signals.}

It is necessary to distinguish between global and local error signals. Local errors signals are the recursively computed signals $\delta_j$. The global error is the derivative of the network's error function with respect to the activity of the output layer. 

\begin{defn}[influence]\label{d:influence}
	The influence of node $j$ on node $k$ is $\tau_{jk}:=  w_{jk} \bbO_k$. The \textbf{influence} of node $j$ on the next layer is $\tau_j := \sum_{\{k|j\rightarrow k\}}\tau_{jk}$. The \textbf{total influence} of node $j$ on downstream nodes is
	\begin{equation}
	\label{e:err_sig}
		\pi_j 
		:= \left(\sum_{\{k|j\rightarrow k\}}\tau_{jk}\left(\sum_{\{l|k\rightarrow l\}}\tau_{kl}\left(\sum_{\{m|l\rightarrow m\}}\cdots\right)\right)\right),
	\end{equation}
	the sum over all paths from $j$ to the output layer.
\end{defn}

Our third result is that Backprop's error signals factorize whenever a neural network has 1-dimensional outputs. 

\begin{thm}[error signal factorization]\label{t:factorization}
	 Suppose neural network $N$ has scalar output and let $\beta=\frac{\partial E}{\partial x_o}$ be the global error. Then, the error signal of a hidden node $j$ factorizes as 
	 \begin{equation}
 		\label{e:rlbp}
	 	\delta_j = \beta\cdot \pi_j = \big(\text{global error}\big)\cdot\big(\text{total influence}_j\big).
	 \end{equation}
\end{thm}

The theorem holds in the setting of nonparametric regression. Multi-label classification is excluded.

\begin{proofsketch}
	Backprop recursively updates weight vectors using the chain rule, recall \eqref{e:bp}. When the output is one-dimensional, $x_o$ contributes $\beta$ to the recursive computation of $\pi_j$ over hidden nodes.
\end{proofsketch}

\paragraph{Kickback.}
We are now ready to introduce Kickback.

\begin{algol}[Kickback]\label{a:kickback}
	The \textbf{truncated feedback} $\epsilon_j$ at node $j$ is
	\begin{equation}
		\label{e:trunc_err_sig}
		\epsilon_j := \beta\cdot \tau_j 
		= \big(\text{global error}\big)\cdot \big(\text{influence}_j\big).		
	\end{equation}
	Under \textbf{Kickback}, hidden nodes perform gradient descent on the rectilinear loss with truncated feedback:	
	\begin{equation}
		\label{e:kb_w}
		\Delta w_{ij} 
		\propto -\nabla_{w_{ij}} \bell_{RL}(\wt_j,\x,\epsilon_j) 
		= -\beta\cdot \tau_j \cdot x_i\cdot \bbO_j.
	\end{equation}
\end{algol}

Kickback and Backprop are contrasted in Figure~1 and in equations \eqref{e:rlbp} versus \eqref{e:trunc_err_sig}. Importantly, Kickback eliminates the need for nodes to communicate error signals -- as distinct from their outputs.

\paragraph{Kickback as time-averaged Backprop.}
Truncating the feedback signal, from \eqref{e:rlbp} to \eqref{e:trunc_err_sig}, preserves more information than appears at first glance. The truncated signal received by node $j$ \emph{explicitly} depends on $j$'s influence on the next layer. However, Kickback \emph{implicitly} incorporates information about the influence of multiple layers.

For simplicity, suppose there is no regularizer and that the learning rate $\eta$ is constant. Then, summing over the updates in \eqref{e:grad1}, a weight at time $T$ is $w_{ij}^T = \eta \sum_{t \in F_j} \varphi^t_j x_i^t$. In the specific case of Kickback, the weight is
\begin{align*}
	\label{e:slow-average1}
	w_{ij}^T  = \eta \sum_{t \in F_j} \big(\beta^t x_i^t \tau_j^t\big)
	 = \eta \sum_{t\in F_j} \Big(\beta^t x_i^t 
	\sum_{\{k|j\rightarrow k\}}w_{jk}^t\bbO_k\Big).
\end{align*}
The weight $w_{ij}^T$ thus implicitly incorporates the effect of interactions $\tau_{jk}^t=w_{jk}^t\bbO_k^t$ in the next layer down, and so on recursively.

\paragraph{Coherence.}
With a small enough learning rate, gradient descent will tend towards a local minimum. Kickback does not perform gradient descent on the error function since it uses modified feedback signals. Thus, without further assumptions, it is not guaranteed to improve performance. Our fourth result is to provide a sufficient condition.

\begin{defn}[coherence]\label{d:coherence}
	Node $j$ is \textbf{coherent} when $\tau_j>0$. A network is coherent when all its nodes are coherent.
\end{defn}

\begin{eg}[signed coherence]\label{eg:signed}
	An easy way to guarantee coherence for every node is to impose the purely \emph{local} condition that all connections targeting positive nodes have positive weights, and similarly that all connections targeting negative nodes have negative weights. 
\end{eg}	

If a network is coherent, then increasing a positive node's firing rate increases the average (signed) activity in the next layer and \emph{all downstream layers}. Increasing the activity of negative nodes has the opposite effect.

On the other hand, if a network is not coherent, then nothing can be said in general about how the activity of nodes in one layer affects other layers. 

Coherence thus enforces \emph{interpretability}: it ensures that a node's influence on the next layer is indicative of its \emph{total} influence on all downstream layers. 

\begin{thm}[coherence $\implies$ Kickback reduces error]\label{t:coherence}
	If a network is coherent then weight updates under Kickback, with a sufficiently small learning rate, improve performance.
\end{thm}

\begin{proofsketch}
	It suffices to show that the feedback has the same sign under Backprop, $\delta_j = \beta\cdot \pi_j$, and Kickback, $\epsilon_j=\beta\cdot \tau_j$ for an arbitrary hidden node $j$. 

	If $j$ is coherent then $\tau_j>0$. If, furthermore, all downstream nodes are coherent, then unraveling \eqref{e:err_sig} obtains that $\pi_j>0$. The result follows.
\end{proofsketch}

Under Backprop, each node's total influence is computed explicitly. Kickback makes do with less information: a node ``knows'' its influence on the next layer, but does not ``know'' its total influence.

\section{Biological relevance}
\label{s:selectron}

There is a direct link from Kickback to neurobiology provided by the \emph{selectron}: a simplified model neuron \cite{bb:12}. The selectron is derived from standard models of neural dynamics and learning -- the Spike Response Model (SRM) and Spike-Timing Dependent Plasticity (STDP) --  by taking the so-called ``fast-time constant limit'' to go from continuous to discrete time. 

\begin{thm}[selectron]\label{t:stdp}
	The fast time-constant limit of the SRM \cite{gerstner:02} is a node that outputs $1$ if $\langle\wt,\x\rangle>0$ and $0$ otherwise. 

	Weight updates in the fast time-constant limit of neuromodulated STDP \cite{song:00} are
	\begin{equation}
		\label{e:stdp}
		\Delta w_{ij} \propto \nu\cdot x_i \cdot \bbO_j = 
		\begin{cases}
			\nu\cdot x_i & \text{if }\langle\wt_j,\x\rangle >0 \\
			0 & \text{else,}
		\end{cases}		
	\end{equation}
	where $\nu$ is a global, scalar-valued neuromodulatory signal. 

	The weight updates in \eqref{e:stdp} are gradient \emph{ascent} on
	\begin{equation*}
		\label{e:reward}
		\text{Reward}(\wt,\x,\nu) 
		:= \nu\cdot P_\wt(\x)
		= \begin{cases}
			\nu\langle\wt,\x\rangle & \text{if }\langle\wt_j,\x\rangle >0 \\
			0 & \text{else.}
		\end{cases}
	\end{equation*}
\end{thm}

Setting $\varphi := -\nu$ in $\text{\emph{Reward}}(\wt,\x,\nu)$ recovers the rectilinear loss in Definition~\ref{d:rl}. The selectron thus maximizes a \emph{rectilinear reward} via the same weight updates used to minimize the rectilinear loss. The difference between the two models is that the selectron has 0/1-valued outputs (spikes), whereas nodes have real-valued outputs (firing rates).

\begin{proof}
	\cite{bb:12}.
\end{proof}

Kickback's weight updates are $\Delta w_{ij}\propto -\beta\cdot \tau_j\cdot x_i\cdot\bbO_j$. Each factor has a biological analog. The global error, $\beta$, corresponds to neuromodulators, such as dopamine, that have been experimentally observed to signal prediction errors for future rewards \cite{schultz:97}. The \emph{kickback} term, $\tau_j$, corresponds to NMDA backconnections that have a multiplicative effect on synaptic updates, proportional to the weighted sum of downstream activity \cite{vargas:03,roelfsema:05}. The feedforward term, $x_i$, corresponds to pre-synaptic spiking activity \cite{song:00}. Finally, the signed indicator function $\bbO_j$, ensures that only active nodes update their weights -- thereby playing the role of post-synaptic activity in STDP.

The regret bound in Theorem~\ref{t:regret} is also biologically significant. Synapses incur a significant metabolic cost \cite{tononi:14}. Regularizing synaptic weights provides a way to quantify metabolic costs. Indeed, limits on the physical size and metabolic budget of synapses suggest that synaptic weights may be constrained to an $\ell_1$-ball \cite{bb:12}. 

To the best of our knowledge, Theorem~\ref{t:regret} is the first \emph{adversarial} generalization bound for a biologically derived model. The generalization bound for the selectron in \cite{bb:12} assumes that inputs are \emph{i.i.d}. Moving beyond the \emph{i.i.d.} assumption is important because biological organisms face adversarial environments. 

The final ingredient is coherence. Investigating biologically plausible mechanisms that ensure coherence (or some other sufficient condition) is deferred to future work.

\section{Experiments}
\label{s:experiments}

\paragraph{Goals.}
Our primary aim is to compare Kickback's performance to Backprop. We present results on two robotics datasets, SARCOS\footnote{Taken from \texttt{www.gaussianprocess.org/gpml/data/}.} and Barrett WAM\footnote{Taken from \texttt{http://www.ias.tu-darmstadt.de/\\Miscellaneous/Miscellaneous}.}. Kickback's performance across multiple hidden layers is of particular interest, since it truncates errors. Results for 3 hidden layers are reported; results for 1 and 2 hidden layers were similar.\footnote{In short: the performance of both Kickback and Backprop is worse, but still comparable, with fewer layers.} A secondary aim is to investigate the effect of coherence. 

Competing on the datasets tackled by deep learning algorithms is not yet feasible. Further work is required to adapt Kickback to multiclass learning.

\begin{figure*}[t]
	\centering	
	\vspace{-6mm}
	\subfigure[Barrett 2]{\includegraphics[width=2.27in]{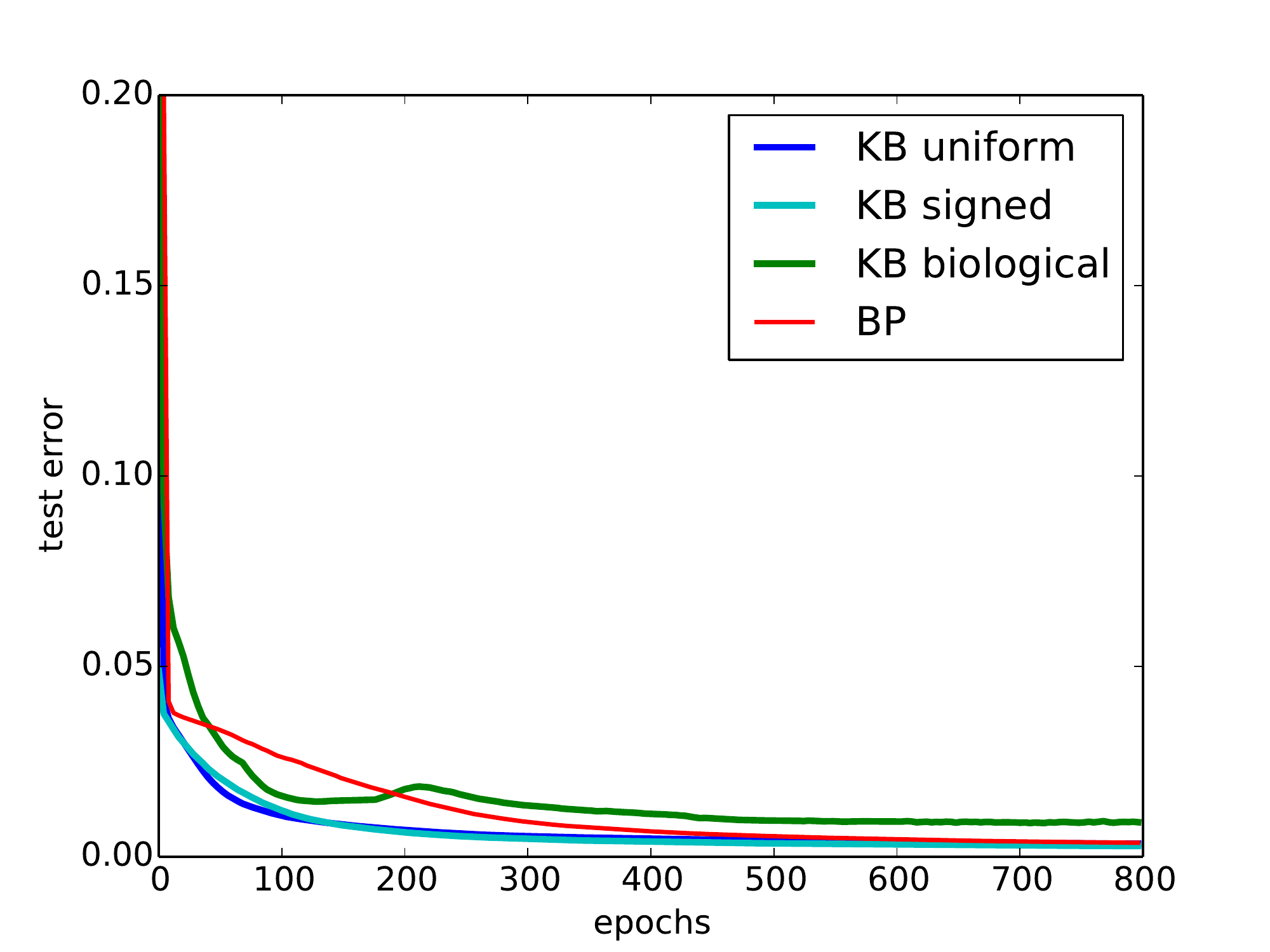}
	\label{f:b2}}
	\subfigure[Barrett 3]{\includegraphics[width=2.27in]{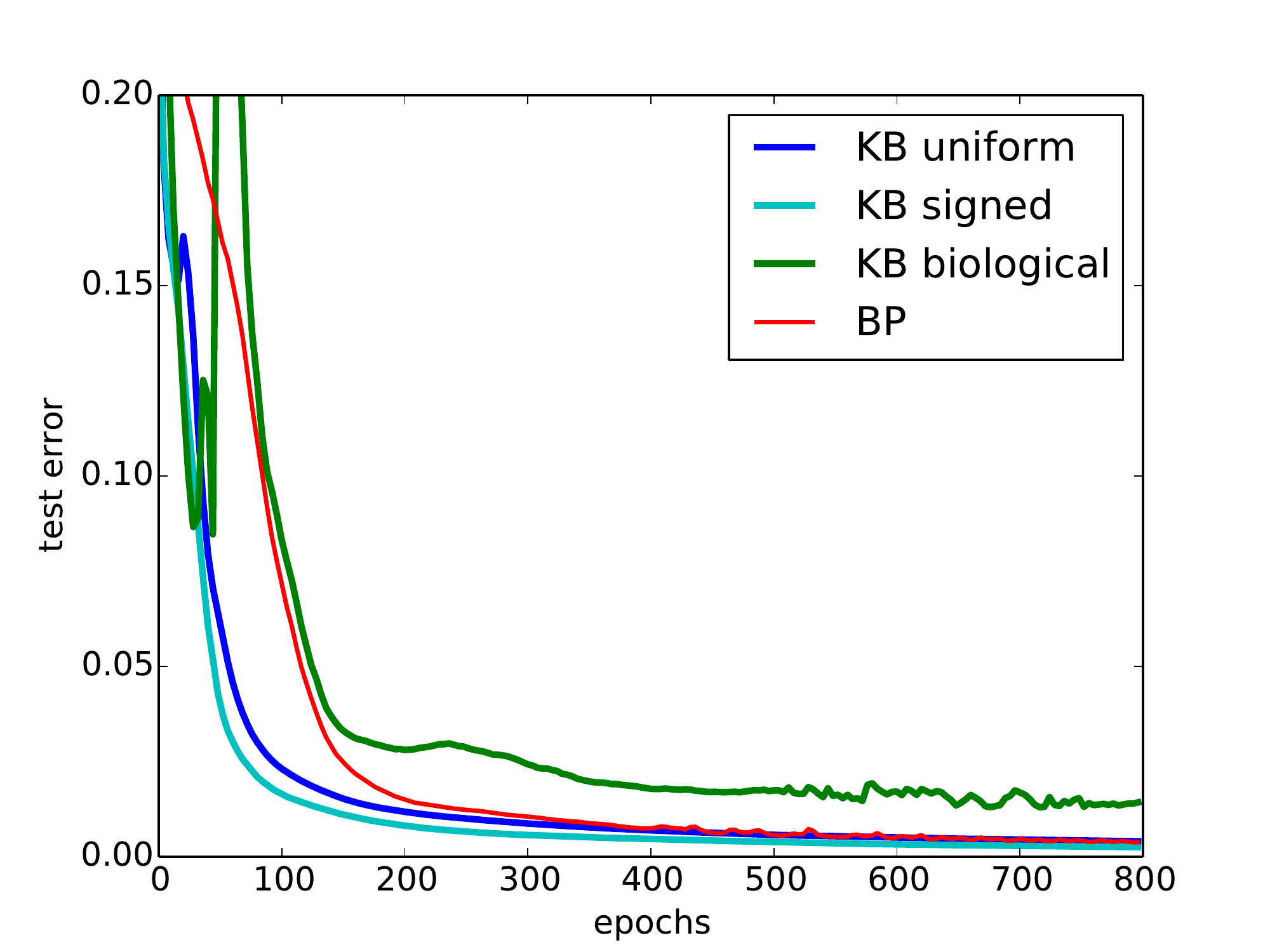}
	\label{f:b3}}
	\subfigure[Barrett 4]{\includegraphics[width=2.27in]{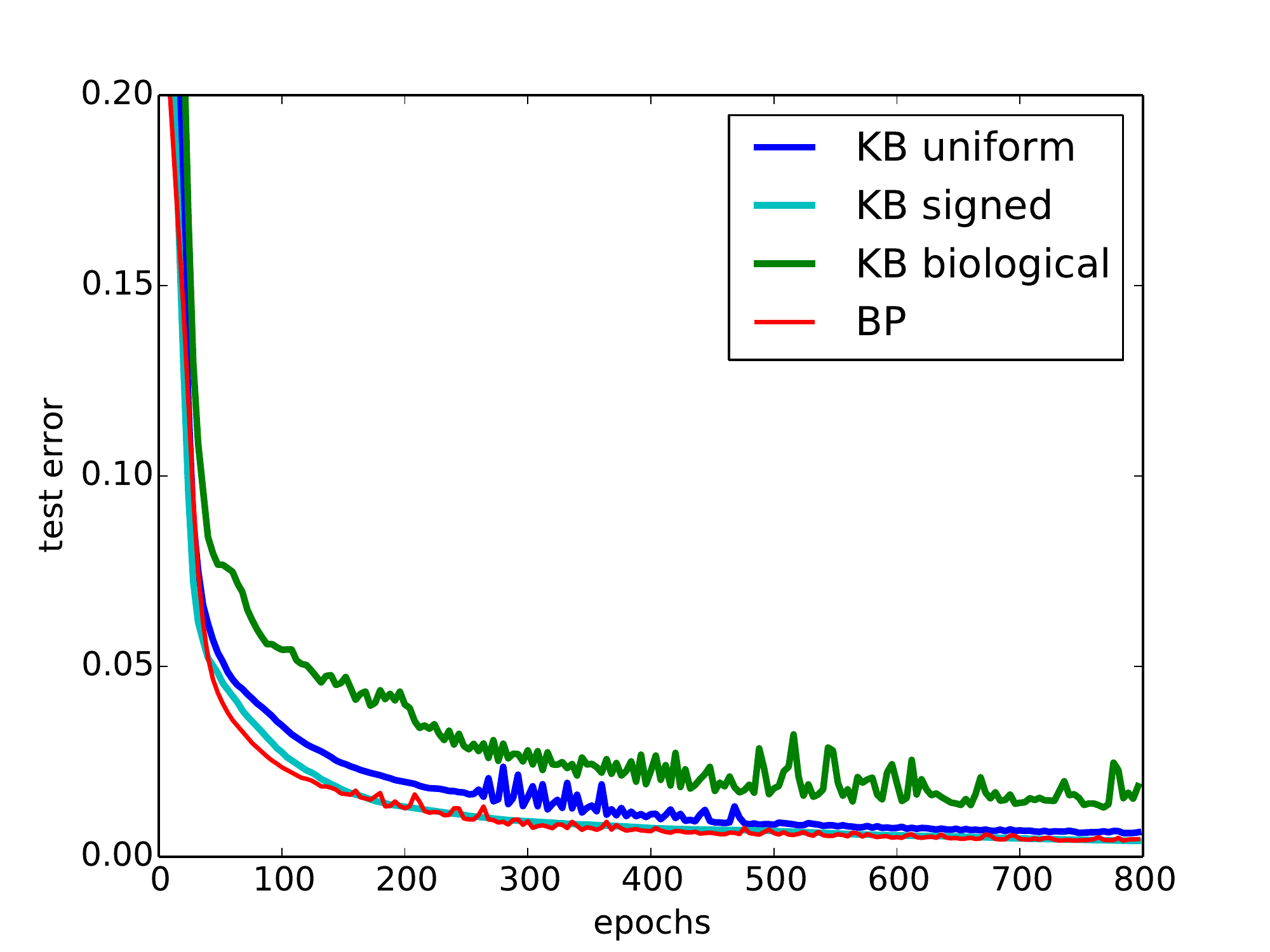}
	\label{f:b4}}	
	\vspace{-4mm}
	\subfigure[SARCOS 1] {\includegraphics[width=2.27in]{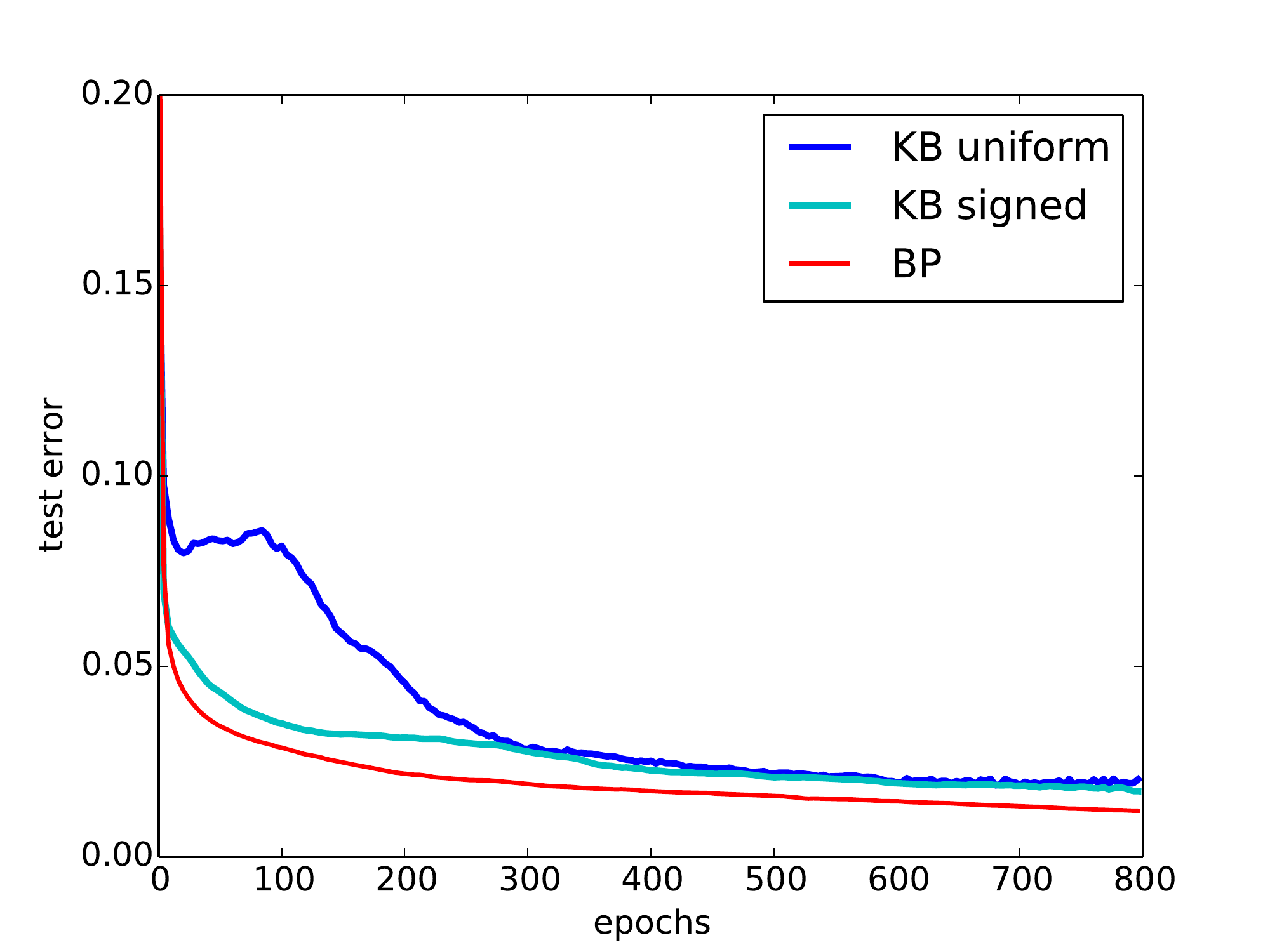}
	\label{f:s1}}
	\subfigure[SARCOS 3] {\includegraphics[width=2.27in]{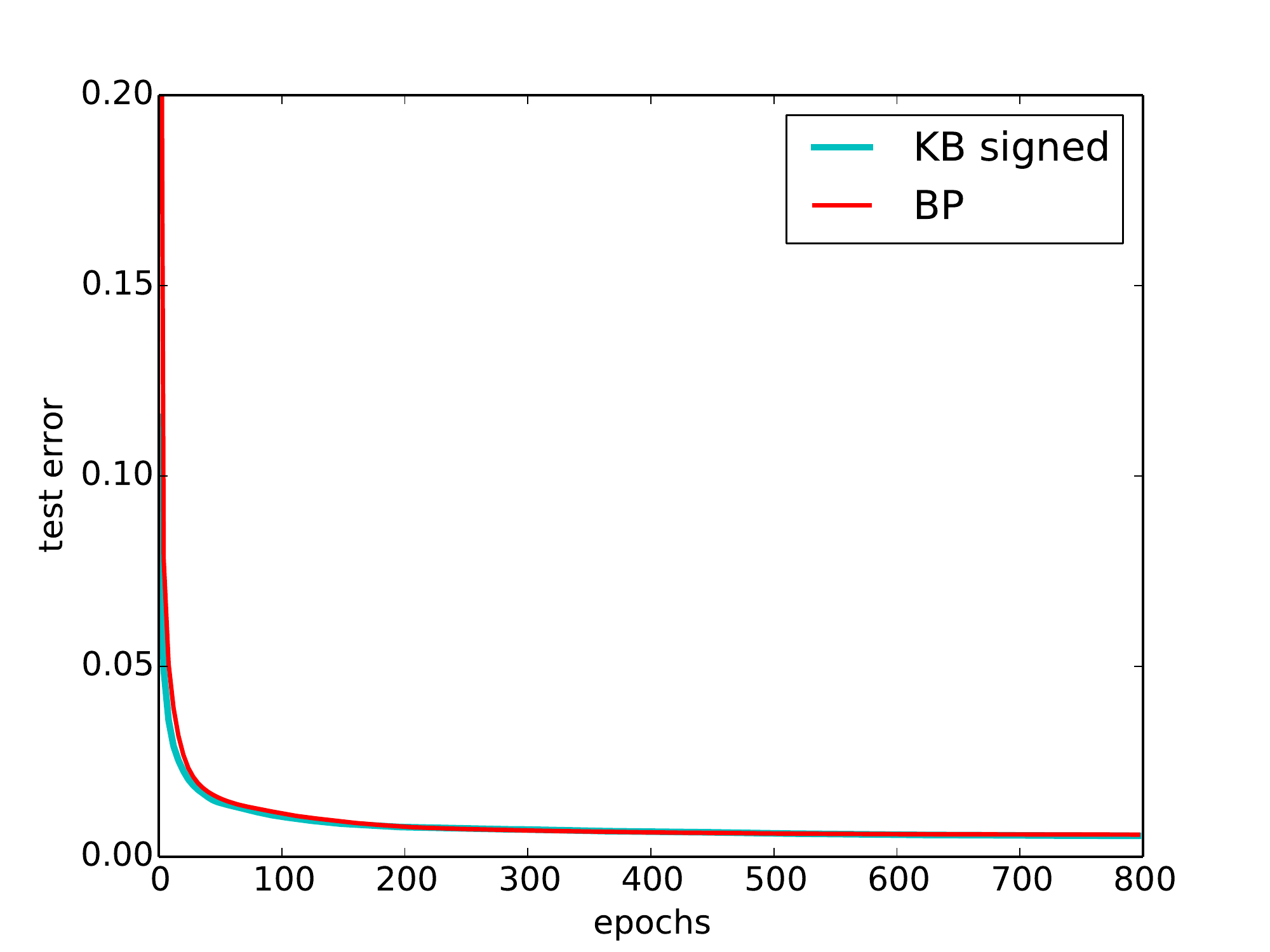}
	\label{f:s3}}
	\subfigure[SARCOS 7] {\includegraphics[width=2.27in]{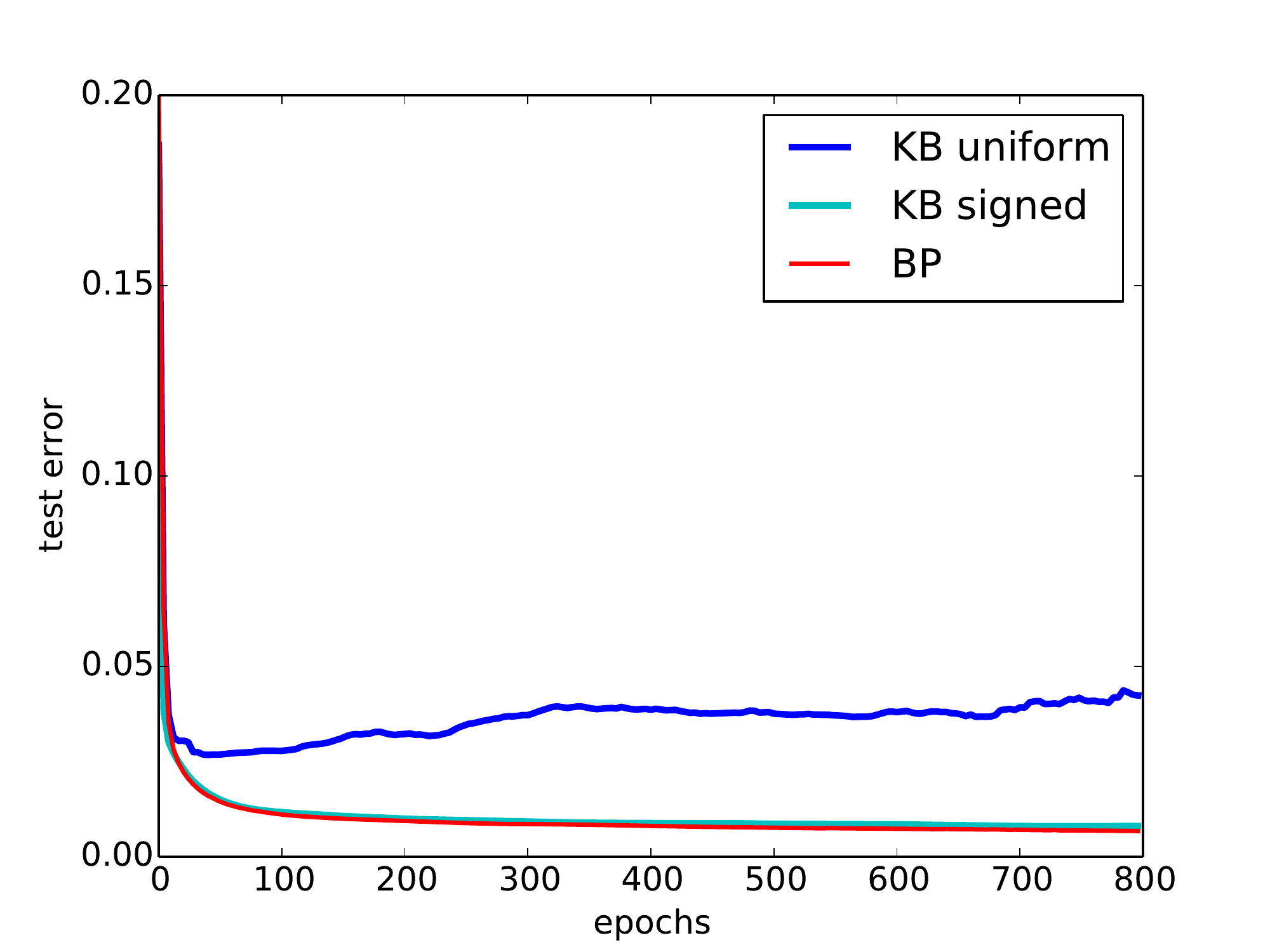}
	\label{f:s7}}
	\caption{\textbf{Mean-squared test error per epoch for Kickback and Backprop.}}
\end{figure*}

\paragraph{Architecture.}
Experiments were performed on a 5-layer network with 2 output nodes, 10, 100 and 200 nodes in three hidden layers, and with the input layer directly drawn from the data.  Experiments were implemented in Theano \cite{bergstra:10}. All nodes are rectifiers. We set half of nodes as positive and half as negative. Output nodes perform rectilinear regression, see below, whereas hidden nodes minimize the rectilinear loss on feedback implementing either Kickback or Backprop.

Training was performed in batch sizes of 20. Lower batch-sizes yield better performance at the cost of longer training times. We chose 20 as a reasonable compromise.

\paragraph{Rectilinear regression.}
Recently, \cite{glorot:11} introduced an $\ell_1$ penalty on firing rates, which encourages sparsity and can improve performance. Here, we consider an $\ell_2$-penalty: $\bell_{RL}(\wt,\x,\varphi) - \frac{1}{2}S_\wt(\x)^2$. Weight updates under gradient descent are
\begin{equation}
	\label{e:grad2}
	\Delta\wt \propto 
	\begin{cases}
		(\varphi - \langle\wt,\x\rangle)\x & \text{if }\langle\wt,\x\rangle > 0\\
		0 & \text{else.}
	\end{cases}
\end{equation}
Notice that the penalty $\langle\wt,\x\rangle$ in \eqref{e:grad2} is the firing rate. Comparing with the gradient $(\varphi - \langle \wt,\x\rangle)\x$ of the mean-squared error $\frac{1}{2}(\varphi-\langle\wt,\x\rangle)^2$ shows that the $\ell_2$-activation penalty leads nodes to \emph{perform linear regression on the inputs that cause them to fire} \cite{balduzzi:13mv}. A regret bound analogous to Theorem~\ref{t:regret} holds for rectilinear regression, with a faster convergence rate of $O(\frac{\log |F|}{|F|})$.

Training error is the MSE of the output node with the correct sign\footnote{Recall there is one positive and one negative output rectifier.}; test error is the sum of the output nodes' MSEs.

\paragraph{Initialization and coherence.}
No pretraining was used. We consider two network initializations. The first is \textbf{uniform}: draw weights uniformly at random from an interval symmetric about 0, without regard to coherence. The second initialization, \textbf{signed} is taken from Example~\ref{eg:signed}: draw weights uniformly, then change their signs so that connections targeting positive nodes have positive weights and conversely for negative nodes. \textbf{Signed} guarantees coherence at initialization. Although it is possible to impose coherence during training, we found that doing so was unnecessary in practice.

Results are plotted under both initializations for Kickback -- excepting Panel (e), where \textbf{uniform} failed to converge. For Backprop, the initialization that yielded the \emph{better} performance is reported.

\paragraph{Results.}
We report normalized mean-squared errors. To directly compare the behavior of the algorithms, we report individual runs. Performance was robust to significant changes in tuning parameters: e.g. changing parameters by $2\times$ increased the MSE on SARCOS 3 from .6\% to .8\%.

Each SARCOS dataset consists of 44,484 training and 4,449 test points; Barrett split as 12,000 and 3,000. Parameters were tuned via grid-search with 5-fold cross-validation. Backprop's only parameter is the learning rate. Kickback was implemented with a learning rate tuned for Backprop. Kickback has two additional parameters that rescale the feedback to hidden layers 1 \& 2. We observed that tuning via cross-validation typically set the rescaling factors such that the truncated errors are rescaled to about same magnitude, on average, as Backprop's feedback. 

Kickback and Backprop are competitive with non-parametric methods such as kernel regression, e.g. \cite{kpotufe:13}. Kickback performs best with \textbf{signed} initialization, as expected from Theorem~\ref{t:coherence}. With \textbf{signed} initialization, Kickback almost exactly matches Backprop in all 6 datasets. Importantly, Kickback continues to reduce the MSE after 100s of epochs; following the correct gradient even when the error is small.

The comparison between Backprop and Kickback is not completely fair: Kickback's additional parameters cause it to outperform Backprop in panel~(b). We have endeavored to keep the comparison as level as possible.

\paragraph{The effect of coherence.}
Kickback's performance was better than expected: coherence was not imposed after initialization under \textbf{signed}; and no guarantees are applicable to \textbf{uniform}. A possible explanation is that Kickback preserves or increases coherence.

To test this hypothesis, we quantified the coherence of layer $\alpha$ as $\textrm{coh}(L_\alpha) = \frac{\sum_{j\in L\alpha}\tau_j}{\sum_{j\in L\alpha}|\tau_j|}$, which lies in $[-1,1]$. With \textbf{signed} initialization, coherence consistently remained above 0.9 under Kickback; but exhibited considerable variability under Backprop. With \textbf{uniform} initialization, Kickback increased the coherence of hidden layers 2 \& 3, from 0 to $>0.5$, with the exception of panel~(c). Backprop did not alter coherence in any consistent way. 

Barrett 4 is the only dataset where nodes become incoherent $(\textrm{coh}<0)$ on average. The oscillations in Panel~(c) for \textbf{uniform} arise because Kickback is not guaranteed to follow the training error gradient in the absence of coherence. It is surprising the network learns at all.  Note that oscillations do not occur when networks are given a \textbf{signed} initialization.

\section{Conclusion}

A necessary step towards understanding how the brain assigns credit is to develop a minimal working model that fits basic constraints. 

Backprop solves the credit assignment problem. It is one of the simplest and most effective methods for learning representations. In combination with various tricks and optimizations, it continues to yield state-of-the-art performance. However, it flouts a basic constraint imposed by neurobiology: it requires that nodes produce error signals that are distinct from their outputs.

Kickback is a stripped-down version of Backprop motivated by theoretical (Theorems~\ref{t:Backprop}--\ref{t:coherence}) and biological (Fig.~1 and Theorem~\ref{t:stdp}) considerations. Under Kickback, nodes perform gradient descent, or ascent, on the representation -- that is, the kicked back activity -- produced by the next layer. The sign of the global error determines whether nodes follow the gradient downwards, or upwards.

Kickback is the first competitive algorithm with biologically plausible credit-assignment. Earlier proposals were not competitive and only worked for one hidden-layer (Kickback works well for $\leq3$ hidden-layers; we have not tested $\geq4$). Kickback's simplified signaling is suited to hardware implementations \cite{indiveri:11a,Nere:2012fk}. 

An important outcome of the paper is a new formulation of Backprop in terms of interacting local learners, that may connect deep learning to recent developments in multi-agent systems \cite{seuken:08,sutton:11} and mechanism design \cite{balduzzi:14cpm}.

Kickback's rescaling factors (1 per hidden layer) are a loose-end that require addressing in future work.

Perhaps the most important direction is to extend Kickback to multiclass learning. For this, it is necessary to consider multidimensional outputs, in which case the derivative of the energy function with respect to the output layer is not a scalar. A natural approach to tackle this setting is to use more sophisticated global error signals. Indeed, modeling the neuromodulatory system as producing scalar outputs is a vast oversimplification \cite{dayan:12}.

Finally, reinforcement learning is a better model of how an agent adapts to its environment than supervised learning \cite{veness:10}. A natural avenue to explore is how Kickback, suitably modified, performs in this setting. 

{
\paragraph{Acknowledgements.}
We thank Jacob Abernethy and Satinder Singh for useful conversations. This research was supported in part by SNSF grant 200021\_137971.
}

{
\small

\bibliographystyle{aaai}
}

\end{document}